\documentclass[11pt,letterpaper]{article}
\usepackage{emnlp2016}
\usepackage{times}
\usepackage{latexsym}

\usepackage{url}
\usepackage{color}
\usepackage{float}
\usepackage{tabularx}
\usepackage{subfig}
\usepackage{multirow}
\usepackage{latexsym}
\usepackage{amsmath}
\usepackage{amssymb}
\usepackage{multirow}
\usepackage{pgfplots}
\usepackage{algorithm}
\usepackage{algorithmic}

\DeclareMathOperator*{\argmax}{arg\,max}
\DeclareMathOperator*{\softmax}{softmax}

\def\bu{\mathbf{u}}

\def\w{\mathbf{w}}

\def\e{\mathbf{e}}

\def\bb{\mathbf{b}}

\def\bs{\mathbf{s}}
\def\bo{\mathbf{o}}
\def\bd{\mathbf{d}}

\def\bW{\mathbf{W}}

\def\R{\mathbf{R}}

\allowdisplaybreaks

\newenvironment{itemize*}%
  {\begin{itemize}%
    \setlength{\itemsep}{0pt}%
    \setlength{\parskip}{0pt}}%
  {\end{itemize}}
  \newenvironment{enumerate*}%
  {\begin{enumerate}%
    \setlength{\itemsep}{0pt}%
    \setlength{\parskip}{0pt}}%
  {\end{enumerate}}
\emnlpfinalcopy
\title{End-to-End Neural Sentence Ordering Using Pointer Network}

\date{}
\author{Jingjing Gong*, Xinchi Chen*, Xipeng Qiu, Xuanjing Huang\\
 Shanghai Key Laboratory of Intelligent Information Processing, Fudan University\\
School of Computer Science, Fudan University\\
825 Zhangheng Road, Shanghai, China\\
\{jjgong15, xinchichen13, xpqiu, xjhuang\}@fudan.edu.cn}

\begin{document}

\maketitle

\begin{abstract}
Sentence ordering is one of important tasks in NLP. Previous works mainly focused on improving its performance by using pair-wise strategy. However, it is nontrivial for pair-wise models to incorporate the contextual sentence information. In addition, error prorogation could be introduced by using the pipeline strategy in pair-wise models. In this paper, we propose an end-to-end neural approach to address the sentence ordering problem, which uses the pointer network (Ptr-Net) to alleviate the error propagation problem and utilize the whole contextual information. Experimental results show the effectiveness of the proposed model. Source codes\footnote{https://github.com/fudannlp} and dataset\footnote{http://nlp.fudan.edu.cn/data/} of this paper are available.
\renewcommand*{\thefootnote}{\fnsymbol{footnote}}
\footnotetext[1]{Jingjing Gong and Xinchi Chen contributed equally to this work.}
\renewcommand*{\thefootnote}{\arabic{footnote}}
\end{abstract}

\section{Introduction}
Recently, sentence ordering task attracts more focus in NLP community \cite{chen2016neural,agrawal2016sort,li2016neural} as its importance on many succeed applications such as multi-document summarization, etc.

The goal of sentence ordering is to arrange a set of sentences into a coherent text in a clear and consistent manner \cite{grosz1995centering,van1999semantic,barzilay2008modeling}.

Most of previous researches of sentence ordering are pair-wise models. \newcite{chen2016neural} and \newcite{agrawal2016sort} proposed pair-wise models followed by a beam search decoder to seek the ground truth sentence order. \newcite{li2016neural} additionally employed the graph based method \cite{lapata2003probabilistic} to rank sentences. However, their methods could be easily affected by the performance of independent sentence pairs without their contextual sentences. 

In this paper, we propose an end-to-end neural sentence ordering approach based on the pointer network (Ptr-Net) \cite{vinyals2015pointer}. Pointer network can deal with some combinatorial optimization problems with attention mechanism, such as sorting the elements of a given set. Our proposed model can take a set of random sorted sentences as input, and generate an ordered sequences. With pointer network, our proposed model could utilize the whole contextual information to specify the sort order. In addition, we further evaluate the robustness of our model by adding unrelated noisy  sentence to the sentence set, which is nontrivial for pair-wise models.
Experimental results on two datasets show that proposed model achieves the state-of-the-art performance even it only works with the greedy decoding.

 The contributions of this paper can be summarized as follows:
 \begin{enumerate*}
   \item Instead of pair-wise sentence ordering, we propose an end-to-end neural approach to generate the order of sentences, which could exploit the contextual information and alleviate the error propagation problem.
   \item We perform extensive empirical experiments and achieve the state-of-the-art performance even it only works with the greedy decoding.
   \item We design a new and harder experiment to evaluate the robustness of our model. We add unrelated noisy sentences to the candidate set, and wish the system can sort the sentences correctly while discarding the noisy sentences.
 \end{enumerate*}

\section{Pointer Network for Neural Sentence Ordering}
\subsection{Task Description}
Sentence ordering task aims to rank a set of sentences in a clear and consistent manner. Specifically, given $n$ sentences $\bs = s_1,s_2,\dots,s_n$, the aim is to find the gold order $\bo^*$ for these sentences:
\begin{equation}
  s_{o^*_1} \succ s_{o^*_2}\succ \dots\succ s_{o^*_{n}},
\end{equation}
which has the maximal probability of given sentences $P(\bo^* | \bs )$:
\begin{equation}
  P(\bo^* | \bs ) > P(\bo | \bs ), \forall \bo \in \Psi,
\end{equation}
where $\bo$ indicates any order of these sentences and $\Psi$ indicates the set of all possible orders.

\subsection{Model Architecture}
\begin{figure}[t]
  \centering
  \includegraphics[width=0.48\textwidth]{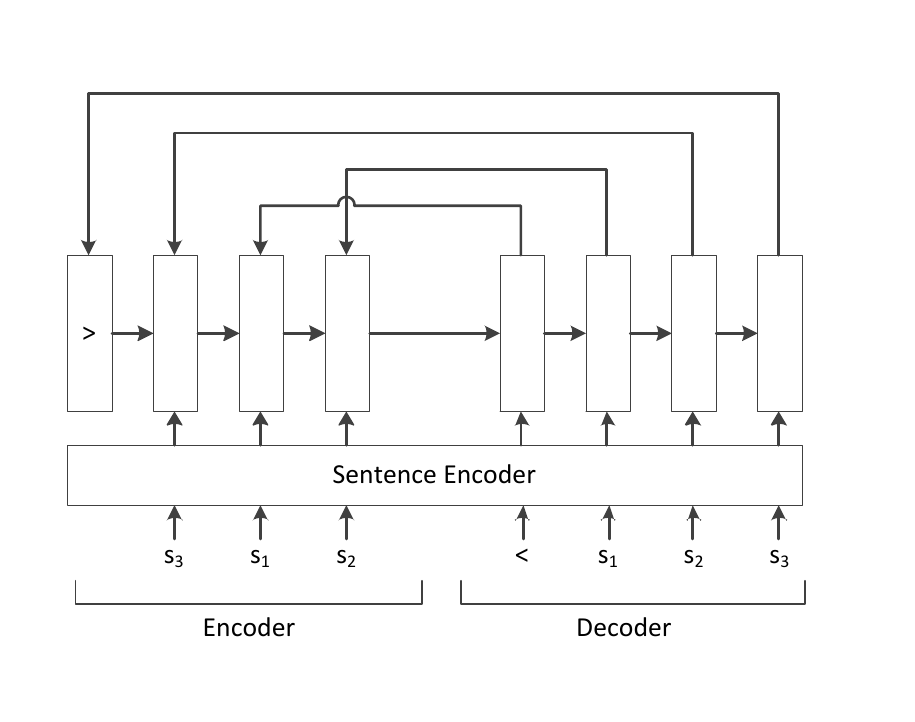}
  \caption{Pointer networks for neural sentence ordering.}\label{fig:ptrnet}
\end{figure}
Previous works mainly focused on pair-wise models \cite{chen2016neural,agrawal2016sort,li2016neural}, which lack contextual information and introduce error propagation for the pipe line strategy. Instead of decomposing the score to independent pairs, in this paper, we propose an end-to-end neural approach based on pointer network (Ptr-Net) to score the entire sequence of sentences.

The model architecture is shown in Figure \ref{fig:ptrnet}. Specifically, the probability of a specific order $\bo$ of given sentences $P(\bo | \bs )$ could be formalized as:
\begin{equation}
  P(\bo | \bs ) = \prod_{i=1}^n P(o_i | o_{i-1},\dots,o_1, \bs).
\end{equation}
The probability $P(o_i | o_{i-1},\dots,o_1, \bs)$ is calculated by Ptr-Net:
\begin{gather}
  P(o_i | o_{i-1},\dots,o_1, \bs) = \softmax(\bu^i)_{o_i}\\ \label{eq:prob_position}
 \bu^i_j = v^\intercal \tanh(\bW^\intercal \left[\begin{array}{c}
                                                \e_j\\
                                                \bd_i
                                                \end{array}\right]
                                                                ), j = (1,\dots, n)
\end{gather}
where $\e_j,\bd_i \in \R^{h}$ are outputs of encoder and decoder of Ptr-Net respectively. $v \in \R^{h}$ and $\bW \in \R^{(2h)\times h}$ are trainable parameters.
\paragraph{Encoder}
The encoding representation of Ptr-Net could be formalized as:
\begin{equation}
  \e_j = \text{LSTM} (\text{Enc}(s_{o_j}), \e_{j-1}), j = (1,\dots, n),
\end{equation}
where Enc$(s_{o_j})$ indicates the encoding of sentence $s_{o_j}$. The sentence encoding function Enc$(\cdot)$ and function LSTM$(\cdot)$ will be further interpreted in Section \ref{sec_sentence_encode}.

Notably, the initial state of encoder is $\e_0 = \mathbf{0}$.
\paragraph{Decoder}
Similarity, the decoding representation of Ptr-Net is formalized as:
\begin{equation}
  \bd_i = \text{LSTM} (\text{Enc}(s_{o_i}), \bd_{i-1}), i = (1,\dots, n).
\end{equation}

Notably, the initial state of decoder is $\bd_0 = \e_n$.

\subsection{Sentence Encoding} \label{sec_sentence_encode}
Since Ptr-Net receives fixed length vectors as inputs, we need firstly encode sentences with variational length. Inspired by \newcite{chen2016neural}, we tried three types of encoders: continues bag of words (CBoW), convolutional neural networks (CNNs) and long short-term (LSTM) neural networks.
\subsubsection{Continues Bag of Words}
Continues bag of words (CBoW) model \cite{mikolov2013efficient} simply averages the embeddings of words of a sentence. Formally, given the embeddings of $n_w$ words of a sentence $s$, $\w_1, \dots, \w_{n_w}$, the sentence embedding Enc$(s)$ is:
\begin{equation}
  \text{Enc}(s) = \frac{1}{n_w}\sum_{k=1}^{n_w} \w_k,
\end{equation}
where Enc$(s), \w_k \in \R^{d_e}$. $d_e$ is a hyper-parameter, indicating word embedding size.
\subsubsection{Convolutional Neural Networks}
Convolutional neural networks (CNNs) \cite{simard2003best} are biologically-inspired variants of multiple layer perceptions (MLPs). Formally, sentence $s$ with $n_w$ words could be encoded as:
\begin{align}
    \mathbf{cov}_k &= \phi (\bW_{cov}^\intercal (\oplus_{u = 0} ^ {l_f - 1} \w_{k + u}) + \bb_{cov}), \\
    \text{Enc}(s) &= \max_k \mathbf{cov}_k, \label{eq:max_pooling}
\end{align}
where $\bW_{cov} \in \R^{(d \times l_f) \times d_f}$ and $\bb_{cov} \in \R^{d_f}$ are trainable parameters, and $\phi(\cdot)$ is tanh function. Here, $k = 1, \dots, n_w - l_f + 1$, and $l_f$ and $d_f$ are hyper-parameters indicating the filter length and number of feature maps respectively. Notably, $\max$ operation in Eq (\ref{eq:max_pooling}) is a element-wise operation.

\subsubsection{Long Short-term Neural Networks}
Long short-term (LSTM) neural networks \cite{hochreiter1997long} are advanced recurrent neural networks (RNNs), which alleviate the problems of gradient vanishment and explosion. Formally, LSTM has memory cells $\mathbf{c} \in \R^{d_r}$ controlled by three kinds of gates: input gate $\mathbf{i} \in \R^{d_r}$, forget gate $\mathbf{f} \in \R^{d_r}$ and output gate $\mathbf{o} \in \R^{d_r}$:
\begin{align}
\left[ \begin{array}{c}
            \mathbf{i}_t \\
            \mathbf{o}_t \\
            \mathbf{f}_t \\
            \tilde{\mathbf{c}}_t
            \end{array}\right] &= \left[\begin{array}{c}
                                            \sigma\\
                                            \sigma\\
                                            \sigma\\
                                            \phi
                                        \end{array}\right]
                                                 \left( {{\bW}_g}^\intercal
                                                                \left[\begin{array}{c}
                                                                        \w_t\\
                                                                        {\mathbf{h}}_{t-1}
                                                                \end{array}\right]
                                                        + {\bb}_g
                                                \right), \\
               \mathbf{c}_t    &= \mathbf{c}_{t - 1} \odot \mathbf{f}_t + \tilde{\mathbf{c}}_t \odot \mathbf{i}_t, \\
               {\mathbf{h}}_t &= \mathbf{o}_t \odot \phi( \mathbf{c}_t ),
\end{align}
where ${\bW}_g \in \R^{(d + d_r) \times 4d_r}$ and ${\bb}_g \in \R^{4d_r}$ are trainable parameters. $d_r$ is a hyper-parameter indicating the cell unit size as well as gate unit size.
$\sigma(\cdot)$ is sigmoid function and $\phi(\cdot)$ is tanh function.
Here, $t = 1, \dots, n_w$.
Thus, we would represent sentence $s$ as:
\begin{equation}
  \text{Enc}(s) = {\mathbf{h}}_{n_w}.
\end{equation}
\subsection{Order Prediction}
Given $P(\bo|\bs)$, the predicted order $\hat{\bo}$ is the one with highest probability:
\begin{equation}
  \hat{\bo} = \argmax_\bo P(\bo|\bs).
\end{equation}

Since the decoding process, to find $\hat{\bo}$, is a NP hard problem. Instead, we use two strategies to decode a sub optimal result: greedy decoding and beam search decoding.
\paragraph{Greedy Decoding}
At decoding phase of Prt-Net, greedy strategy determines $\hat{\bo} = \hat{o}_1,\dots,\hat{o}_n$ step by step as:
\begin{equation}
  \hat{o}_i = \argmax_{o_i} P(o_i | \hat{o}_{i-1},\dots,\hat{o}_1, \bs).
\end{equation}
\paragraph{Beam Search Decoding}
Beam search strategy always keeps top $b$ terms as candidates each step. Formally, at step $t$, each candidate $\hat{\bo}_1^t = \hat{o}_1, \dots, \hat{o}_t$ has a probability:
\begin{equation}
  P(\hat{\bo}_1^t | \bs) = \prod_{i=1}^t P(\hat{o}_i | \hat{o}_{i-1},\dots,\hat{o}_1, \bs),
\end{equation}
and $b$ candidates with higher probabilities will be kept at $t$ step in the beam.
\section{Training}
Assuming that we have m training examples $(x_i,y_i)_{i=1}^m$, where $x_i$ indicates a sequence of sentences with a specific permutation of $y_i$, and $y_i$ is in gold order $\bo^*$. For obtaining more training data, we randomly generate new permutation for $x_i$ at each epoch. The goal is to minimize the loss function $J(\theta)$:
\begin{equation}
  J(\theta) = - \frac{1}{m} \sum_{i=1}^m \log P(y_i|x_i;\theta) + \frac{\lambda}{2}\|\theta\|_2^2, \label{eq:loss_func}
\end{equation}
where $P(y_i|x_i;\theta) = P(\bo^* | \bs = x_i;\theta)$ and $\lambda$ is a hyper-parameter of regularization term. $\theta$ indicates all trainable parameters.

In addition, we use AdaGrad \cite{duchi2011adaptive} with shuffled mini-batch to train our model. We also use pre-trained embeddings \cite{turian2010word} as initialization. 

\begin{table}[t]
\centering
\begin{tabular}{|c|c|c|c|c|}
 \hline
   \multicolumn{2}{|c|}{Attributes} & $\text{N}$ & $\text{S}_{\text{Avg}}$& $\text{W}_{\text{Avg}}$\\
 \hline
 \multirow{3}*{arXiv}&Train&884,912&5.38&134.58\\
     &Dev&110,614&5.39&134.80\\
    &Test&110,615&5.37&134.58\\
 \hline
 \multirow{3}*{SIND}&Train&40,155&\multirow{3}*{5}&56.71\\
     &Dev&4,990&&57.49\\
    &Test&5,055&&56.31\\
 \hline
 \end{tabular}
\caption{Details of datasets. $\text{N}$ indicates the number of texts. $\text{S}_{\text{Avg}}$ is the average sentence number per text. $\text{W}_{\text{Avg}}$ is the average word count per text.}\label{tab:info_dataset}
\end{table}

\begin{table}[t] \setlength{\tabcolsep}{3pt}
\centering
\begin{tabular}{|c|c|}
  \hline
  Initial learning rate &$\alpha = 0.5$\\
  Regularization&$\lambda = 10^{-5}$\\
  Hidden layer size of Ptr-Net&$h=200$\\
  Filter length of CNN&$l_f = 3,4,5$\\
  Number of feature maps &$d_f = 128$\\
  Hidden size of LSTM& $d_r = 200$\\
  Size of embedding &$d_e = 100$\\
  Beam size & $b=64$\\
    Batch size&$128$\\
  \hline
\end{tabular}
\caption{Hyper-parameter configurations.}\label{tab:paramSet}
\end{table}

\begin{table*}[t]
\centering
\begin{tabular}{|l|*{6}{c|}}
 \hline
 \multirow{2}*{Models}    &   \multicolumn{3}{c|}{arXiv}   &   \multicolumn{3}{c|}{SIND}   \\
 \cline{2-7}
  &PM&LSR&PMR&PM&LSR&PMR\\
 \hline
 \cite{chen2016neural}&82.97&-&33.43&-&-&-\\
 \cite{agrawal2016sort}&-&-&-&73.20&-&-\\
 \hline
 \hline
 CBoW+Ptr-Net&81.93&78.47&33.09&70.69&69.69&09.70\\
 CNN+Ptr-Net &85.00&81.49&38.62&72.17&71.03&11.08\\
 LSTM+Ptr-Net&85.62&82.15&40.00&74.10&72.45&\textbf{12.36}\\
  \hline
  \textbf{+beam search}&&&&&&\\
CBoW+Ptr-Net&82.10	&78.69	&33.43	&72.77	&71.45	&11.45\\
CNN+Ptr-Net&85.26	&81.85	&39.28	&\textbf{74.21}	&72.39	&12.32\\
LSTM+Ptr-Net&\textbf{85.79}	&\textbf{82.40}	&\textbf{40.44}	&74.17	&\textbf{72.52}	&12.34\\
  \hline
 \end{tabular}
\caption{Performances of different models on test sets of datasets. Since there is no noisy sentence here, we constrain the number of output sentences as the same as input, and three scores of PM and LSR are the same in this table.}\label{tab:res_all}
\end{table*}

\section{Experiments}
\subsection{Datasets}
To evaluate proposed model, we adopt two datasets: abstracts on arXiv \cite{chen2016neural} and SIND (Sequential Image Narrative Dataset) \cite{ferraro2016visual}. Examples in SIND dataset all contain 5 sentences, and we only use captions in this paper. The details of two datasets are shown in Table \ref{tab:info_dataset}.
\subsection{Hyper-parameters}
Table \ref{tab:paramSet} gives the details of hyper-parameter configurations. The CNN sentence encoder uses three different filter lengths as \newcite{kim2015character}.
\subsection{Metrics}
To evaluate our model, we use three different metrics: (1) Pairwise metrics; (2) Longest sequence ratio; (3) Perfect match ratio.
\paragraph{Pairwise Metrics}
Pairwise metrics (PM) is the fraction of pairs of sentences whose predicted relative order is the same as the ground truth order (higher is better). Formally, pairwise metrics can be denoted as three scores: precision $P$, recall $R$ and $F$-value.
\begin{align}
  P &= \frac{1}{m}\sum_{i=1}^m\frac{|S(\hat{\bo}_i)\bigcap S(\bo^*_i)|}{|S(\hat{\bo}_i)|},\\
  R &= \frac{1}{m}\sum_{i=1}^m\frac{|S(\hat{\bo}_i)\bigcap S(\bo^*_i)|}{|S(\bo^*_i)|},\\
  F &= \frac{2*P*R}{P + R},
\end{align}
where function $S(\cdot)$ denotes the set of all skip bigram sentence pairs of a text, and function $|\cdot|$ indicates the size of set.

Concretely, we take $\{\hat{\bo} = (2,3,1,4), \bo^* = (1,3,4)\}$ as an example, where 2nd sentence is a noisy term. The pairwise scores of this example are: $P = 1/6, R = 1/3, F =  2/9$.

\paragraph{Longest Sequence Ratio}
Longest sequence ratio (LSR) calculates the radio of longest correct sub-sequence (consecutiveness is not necessary, and higher is better). Formally, LSR can be denoted as three scores: precision $P$, recall $R$ and $F$-value.
\begin{align}
  P &= \frac{1}{m}\sum_{i=1}^m \frac{|L(\hat{\bo}_i, \bo^*_i)|}{|\hat{\bo}_i|},\\
  R &= \frac{1}{m}\sum_{i=1}^m \frac{|L(\hat{\bo}_i, \bo^*_i)|}{|\bo^*_i|},\\
  F &= \frac{2*P*R}{P + R},
\end{align}
where function $L(\cdot)$ denotes the number of elements in longest correct sub-sequence. The value of function $L(\hat{\bo} = (2,3,1,4), \bo^* = (1,3,4))$ of the example above is 2.

\paragraph{Perfect Match Ratio}
Perfect match ratio (PMR) calculates the radio of exactly matching case (higher is better):
\begin{equation}
  \text{PMR} = \frac{1}{m}\sum_{i=1}^m \textbf{1}\{\hat{\bo}_i = \bo^*_i\}
\end{equation}

\begin{figure*}[t]\small
  \centering
  \pgfplotsset{width=0.33\textwidth}
\subfloat[PM]{
  \begin{tikzpicture}
    \begin{axis}[
    xlabel={\# of sentences},
    ymax=100,
    legend entries={CBoW,CNN,LSTM,Chen et al.},
    mark size=0.8pt,
    ymajorgrids=true,
    grid style=dashed,
    legend pos= north east,
    legend style={font=\tiny,line width=.5pt,mark size=.2pt,
            /tikz/every even column/.append style={column sep=0.5em}},
            smooth,
    ]
    \addplot [green,dashed,mark=star] table [x index=0, y index=1] {PM_length_analyse.txt};
    \addplot [blue,dashed,mark=square*] table [x index=0, y index=2] {PM_length_analyse.txt};
    \addplot [red,dashed,mark=otimes*] table [x index=0, y index=3] {PM_length_analyse.txt};
    \addplot [black,mark=square] table [x index=0, y index=4] {PM_length_analyse.txt};
    \end{axis}
\end{tikzpicture}
}
\hspace{0em}
\subfloat[LSR]{
  \begin{tikzpicture}
    \begin{axis}[
    xlabel={\# of sentences},
    legend entries={CBoW,CNN,LSTM},
    mark size=0.8pt,
    ymajorgrids=true,
    grid style=dashed,
    legend pos= north east,
    legend style={font=\tiny,line width=.5pt,mark size=.2pt,
            /tikz/every even column/.append style={column sep=0.5em}},
            smooth,
    ]
    \addplot [green,dashed,mark=star] table [x index=0, y index=1] {LSR_length_analyse.txt};
    \addplot [blue,dashed,mark=square*] table [x index=0, y index=2] {LSR_length_analyse.txt};
    \addplot [red,dashed,mark=otimes*] table [x index=0, y index=3] {LSR_length_analyse.txt};
    \end{axis}
\end{tikzpicture}
}
\hspace{0em}
\subfloat[PMR]{
  \begin{tikzpicture}
    \begin{axis}[
    xlabel={\# of sentences},
    legend entries={CBoW,CNN,LSTM,Chen et al.},
    mark size=0.8pt,
    ymajorgrids=true,
    grid style=dashed,
    legend pos= north east,
    legend style={font=\tiny,line width=.5pt,mark size=.2pt,
            /tikz/every even column/.append style={column sep=0.5em}},
            smooth,
    ]
    \addplot [green,dashed,mark=star] table [x index=0, y index=1] {EMR_length_analyse.txt};
    \addplot [blue,dashed,mark=square*] table [x index=0, y index=2] {EMR_length_analyse.txt};
    \addplot [red,dashed,mark=otimes*] table [x index=0, y index=3] {EMR_length_analyse.txt};
    \addplot [black,mark=square] table [x index=0, y index=4] {EMR_length_analyse.txt};
    \end{axis}
\end{tikzpicture}
}
\caption{Performances of different sentence encoders on different numbers of sentences on test set of arXiv dataset using greedy decoding strategy.}\label{fig:res_length}
\end{figure*}
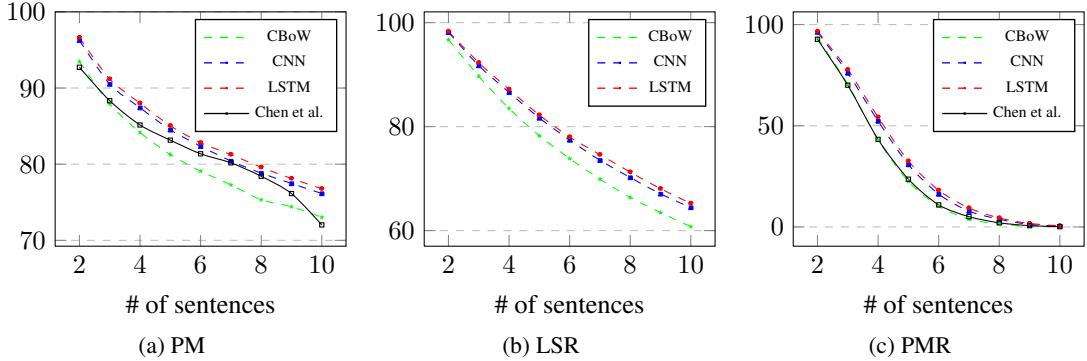

\begin{table}[t] \setlength{\tabcolsep}{3pt}
\centering
\begin{tabular}{|l|*{4}{c|}}
 \hline
 \multirow{2}*{Models}    &   \multicolumn{2}{c|}{arXiv}   &   \multicolumn{2}{c|}{SIND}   \\
 \cline{2-5}
  &Head&Tail&Head&Tail\\
 \hline
 \cite{chen2016neural}&84.85&62.37&-&-\\
 \hline
 \hline
 CBoW+Ptr-Net&85.18&59.93&69.89&45.67\\
 CNN+Ptr-Net &89.59&64.48&72.54&48.80\\
 LSTM+Ptr-Net&\textbf{90.77}&65.80&\textbf{75.24}&52.62\\
  \hline
  \textbf{+beam search}&&&&\\
CBoW+Ptr-Net&84.70	&60.54	&72.61	&51.23\\
CNN+Ptr-Net&89.43	&65.36	&73.53	&53.26\\
LSTM+Ptr-Net&90.47	&\textbf{66.49}	&74.66	&\textbf{53.30}\\
  \hline
 \end{tabular}
\caption{Performances of different models on test sets of datasets.}\label{tab:head_tail}
\end{table}

\begin{table*}[t]\small\setlength{\tabcolsep}{3pt}
\centering
\begin{tabular}{|c|c|c|c|}
\hline
\multicolumn{4}{|c|}{CNN sentence encoder}\\
\hline
2&3&1&1\\
\hline
\multicolumn{1}{|p{0.24\textwidth}|}{
\textbf{
\color{white!24.69!blue}{Our} \color{white!28.50!blue}{second} \color{white!1.62!blue}{question} \color{white!52.89!blue}{regarding} \color{white!55.93!blue}{the} \color{white!56.22!blue}{function} \color{white!58.21!blue}{which} \color{white!36.87!blue}{computes} \color{white!46.58!blue}{minimal} \color{white!14.03!blue}{indices} \color{white!45.62!blue}{is} \color{white!47.51!blue}{whether} \color{white!67.74!blue}{one} \color{white!59.32!blue}{can} \color{white!33.52!blue}{compute} \color{white!51.47!blue}{a} \color{white!45.57!blue}{short} \color{white!28.52!blue}{list} \color{white!53.97!blue}{of} \color{white!58.04!blue}{candidate} \color{white!49.68!blue}{indices} \color{white!58.14!blue}{which} \color{white!47.23!blue}{includes} \color{white!61.05!blue}{a} \color{white!53.49!blue}{minimal} \color{white!35.45!blue}{index} \color{white!54.92!blue}{for} \color{white!58.18!blue}{a} \color{white!49.52!blue}{given} \color{white!0.00!blue}{program}
}
}
&\multicolumn{1}{p{0.24\textwidth}|}{
\textbf{
\color{white!53.23!blue}{We} \color{white!28.44!blue}{give} \color{white!55.87!blue}{some} \color{white!66.35!blue}{negative} \color{white!69.82!blue}{results} \color{white!70.71!blue}{and} \color{white!74.45!blue}{leave} \color{white!76.94!blue}{the} \color{white!72.66!blue}{possibility} \color{white!74.41!blue}{of} \color{white!72.13!blue}{positive} \color{white!71.25!blue}{results} \color{white!74.22!blue}{as} \color{white!70.42!blue}{open} \color{white!49.91!blue}{questions}
}
}
&\multicolumn{1}{p{0.24\textwidth}|}{
\textbf{
\color{white!55.57!blue}{Our} \color{white!50.50!blue}{first} \color{white!63.63!blue}{question} \color{white!73.49!blue}{regarding} \color{white!77.95!blue}{the} \color{white!76.49!blue}{set} \color{white!74.70!blue}{of} \color{white!73.43!blue}{minimal} \color{white!72.70!blue}{indices} \color{white!66.05!blue}{is} \color{white!62.99!blue}{whether} \color{white!69.68!blue}{there} \color{white!75.55!blue}{exists} \color{white!70.75!blue}{an} \color{white!63.38!blue}{algorithm} \color{white!74.57!blue}{which} \color{white!76.73!blue}{can} \color{white!76.33!blue}{correctly} \color{white!75.73!blue}{label} \color{white!67.67!blue}{1} \color{white!52.82!blue}{out} \color{white!65.90!blue}{of} \color{white!63.29!blue}{k} \color{white!66.40!blue}{indices} \color{white!71.97!blue}{as} \color{white!74.72!blue}{either} \color{white!78.29!blue}{minimal} \color{white!78.50!blue}{or} \color{white!76.94!blue}{non} \color{white!77.10!blue}{minimal}
}
}
&\multicolumn{1}{p{0.24\textwidth}|}{
\textbf{
\color{white!25.316177!red}{Our} \color{white!13.904388!red}{first} \color{white!29.020699!red}{question} \color{white!44.284805!red}{regarding} \color{white!64.336563!red}{the} \color{white!67.828491!red}{set} \color{white!51.100006!red}{of} \color{white!62.538704!red}{minimal} \color{white!45.543255!red}{indices} \color{white!48.856609!red}{is} \color{white!42.940521!red}{whether} \color{white!59.580154!red}{there} \color{white!62.920143!red}{exists} \color{white!49.903885!red}{an} \color{white!44.117954!red}{algorithm} \color{white!56.399761!red}{which} \color{white!49.937256!red}{can} \color{white!46.180691!red}{correctly} \color{white!59.332516!red}{label} \color{white!52.417595!red}{1} \color{white!42.085121!red}{out} \color{white!57.623657!red}{of} \color{white!46.471008!red}{k} \color{white!37.604202!red}{indices} \color{white!46.423988!red}{as} \color{white!58.967682!red}{either} \color{white!60.836681!red}{minimal} \color{white!75.768982!red}{or} \color{white!72.482536!red}{non} \color{white!67.921608!red}{minimal}
}
}
\\
\hline
\multicolumn{3}{|c|}{Encoding}&Decoding\\
\hline
\end{tabular}

\begin{tabular}{|c|c|c|c|}
\hline
\multicolumn{4}{|c|}{LSTM sentence encoder}\\
\hline
2&3&1&1\\
\hline
\multicolumn{1}{|p{0.24\textwidth}|}{
\textbf{
\color{white!54.52!blue}{Our} \color{white!37.37!blue}{second} \color{white!22.68!blue}{question} \color{white!53.81!blue}{regarding} \color{white!73.17!blue}{the} \color{white!59.45!blue}{function} \color{white!70.06!blue}{which} \color{white!22.77!blue}{computes} \color{white!67.06!blue}{minimal} \color{white!58.86!blue}{indices} \color{white!72.58!blue}{is} \color{white!65.37!blue}{whether} \color{white!74.82!blue}{one} \color{white!71.78!blue}{can} \color{white!52.72!blue}{compute} \color{white!76.86!blue}{a} \color{white!73.99!blue}{short} \color{white!67.49!blue}{list} \color{white!78.09!blue}{of} \color{white!67.60!blue}{candidate} \color{white!65.58!blue}{indices} \color{white!74.33!blue}{which} \color{white!54.90!blue}{includes} \color{white!77.71!blue}{a} \color{white!66.43!blue}{minimal} \color{white!72.02!blue}{index} \color{white!76.54!blue}{for} \color{white!75.41!blue}{a} \color{white!64.48!blue}{given} \color{white!57.87!blue}{program}
}
}
&\multicolumn{1}{p{0.24\textwidth}|}{
\textbf{
\color{white!73.01!blue}{We} \color{white!69.64!blue}{give} \color{white!73.59!blue}{some} \color{white!67.10!blue}{negative} \color{white!66.96!blue}{results} \color{white!77.05!blue}{and} \color{white!75.93!blue}{leave} \color{white!78.15!blue}{the} \color{white!74.64!blue}{possibility} \color{white!80.00!blue}{of} \color{white!76.93!blue}{positive} \color{white!77.29!blue}{results} \color{white!79.11!blue}{as} \color{white!78.87!blue}{open} \color{white!77.57!blue}{questions}
}
}
&\multicolumn{1}{p{0.24\textwidth}|}{
\textbf{
\color{white!47.14!blue}{Our} \color{white!17.72!blue}{first} \color{white!31.24!blue}{question} \color{white!53.30!blue}{regarding} \color{white!71.11!blue}{the} \color{white!69.85!blue}{set} \color{white!74.24!blue}{of} \color{white!46.37!blue}{minimal} \color{white!28.40!blue}{indices} \color{white!66.78!blue}{is} \color{white!58.81!blue}{whether} \color{white!70.27!blue}{there} \color{white!56.75!blue}{exists} \color{white!73.10!blue}{an} \color{white!33.93!blue}{algorithm} \color{white!72.25!blue}{which} \color{white!73.00!blue}{can} \color{white!68.73!blue}{correctly} \color{white!73.10!blue}{label} \color{white!74.60!blue}{1} \color{white!73.27!blue}{out} \color{white!77.68!blue}{of} \color{white!19.93!blue}{k} \color{white!48.29!blue}{indices} \color{white!76.39!blue}{as} \color{white!74.75!blue}{either} \color{white!71.97!blue}{minimal} \color{white!77.14!blue}{or} \color{white!73.56!blue}{non} \color{white!66.53!blue}{minimal}
}
}
&\multicolumn{1}{p{0.24\textwidth}|}{
\textbf{
\color{white!43.753502!red}{Our} \color{white!10.634781!red}{first} \color{white!0.000000!red}{question} \color{white!40.419090!red}{regarding} \color{white!68.632935!red}{the} \color{white!64.411720!red}{set} \color{white!71.204865!red}{of} \color{white!41.509289!red}{minimal} \color{white!14.274147!red}{indices} \color{white!58.865631!red}{is} \color{white!47.576248!red}{whether} \color{white!64.947464!red}{there} \color{white!42.708298!red}{exists} \color{white!68.286331!red}{an} \color{white!12.504242!red}{algorithm} \color{white!66.540581!red}{which} \color{white!66.248459!red}{can} \color{white!48.981621!red}{correctly} \color{white!66.588715!red}{label} \color{white!69.035049!red}{1} \color{white!65.152229!red}{out} \color{white!74.680252!red}{of} \color{white!10.549225!red}{k} \color{white!34.054546!red}{indices} \color{white!72.908615!red}{as} \color{white!67.563721!red}{either} \color{white!63.807533!red}{minimal} \color{white!72.846664!red}{or} \color{white!65.429451!red}{non} \color{white!64.250679!red}{minimal}
}
}
\\
\hline
\multicolumn{3}{|c|}{Encoding}&Decoding\\
\hline
\end{tabular}
\caption{Case Study. Color indicates importance of words in order prediction. The more important the words are, the darker the color is. In this case, models with CNN and LSTM sentence encoders all make correct predictions.}\label{tab:paragraph_visual}
\end{table*}

\subsection{Results}
Results on different datasets are shown in Table \ref{tab:res_all}. Since there is no noisy sentence here and the number of output sentences is the same as inputs, three scores of PM and LSR are the same in this table. Thus, we only use PM and LSR to denote the same $P,R,F$ values. As we can see, our model outperforms previous works, and achieves the state-of-the-art performance. Our model with LSTM sentence encoder reaches 40.00\% in PMR, which means that 2/5 texts in test set of arXiv dataset are ranked exactly right, 6.57\% boosted compared with work of \newcite{chen2016neural}. With beam search strategy, the performance of our model with LSTM sentence encoder is further boosted to 40.44\% in PRM on test set of arXiv dataset.

Moreover, we investigate the performance of our model when sentence number varies. As \newcite{chen2016neural} did not evaluate their results on LSR metrics, we only compare with them on PM and PMR. As shown in Figure \ref{fig:res_length}, our model with CNN and LSTM sentence encoders outperforms the work of \cite{chen2016neural}. Although the accuracy of our model drops as the number of sentences increases, we could find that our model performs better on text with more sentences compared to pair-wise model.

Since first and last sentences are more special, we also evaluate the accuracy of finding the first and last sentences. As shown in Table \ref{tab:head_tail}, we also achieve significant boost compared to work of \cite{chen2016neural}. As we can see, by using beam search, our model obtains higher accuracy on finding last sentence whereas performance of seeking first sentence drops. However, the performance in total is boosted by using beam search according to the results shown in Table \ref{tab:res_all}, which implies that beam search strategy concentrates more on the entire text.

According to the experimental results above, we could find that Ptr-Net with LSTM sentence encoder almost outperforms the one with CBoW or CNN sentence encoder. Thus, we mainly focus on evaluating our model with LSTM sentence encoder in the rest of this paper.

\subsection{Visualization}
To further understand how our model works, we do some visualizations of Ptr-Net with CNN and LSTM sentence encoders.  As shown in Table \ref{tab:paragraph_visual}, the left side (first three columns) blue terms are input sentences. They are firstly encoded by sentence encoder (CNN or LSTM sentence encoder), then sent to the decoder of Ptr-Net. The right side (last column) red term is the first sentence that our model predicts. After that, this red item is encoded and sent to Prt-Net decoder to generate the 2nd sentence. This visualization shows how important each word is in generating the 2nd sentence. The more important the words are, the darker the color is. As we can see, by using CNN sentence encoder, the model focuses more on individual words, like ``first'', ``second'', which give strong signal for ordering. Contrastly, the model with LSTM sentence encoder trends to find a specific pattern (or compositional features). In this case, the model emphasizes the pattern ``Our X question regarding'', where X could be ordinal numbers like first and second here. However, both of them have some signals like words ``indices'', ``algorithm'', which might be disturbances and hard to interpret.

There still a question remains. How could we determine the importance of each word (the color)? Inspired by the back-propagation strategy \cite{erhan2009visualizing,simonyan2013deep,li2015visualizing,chen2016neural}, which measures how much each input unit contributes to the final decision, we can approximate the importance of words by their first derivatives. Formally, we aim to rank $n$ sentences in gold order $s_1, \dots, s_n$. Assuming the predicted order would be $\hat{\bo}$ and we are predicting the $i$-th sentence $s_{\hat{o}_i}$, the importance of each word $w^j_k$ ($k$-th word in $j$-th sentence $s_{\hat{o}_j}$) is:
\begin{equation}
  A(w^j_k) = \left|\frac{\partial P(\hat{o}_i | \hat{o}_{i-1},\dots,\hat{o}_1,\bs)}{\partial\w^j_k}\right|,
\end{equation}
where $\w^j_k$ is word embedding of word $w^j_k$. Function $|\cdot|$ is the second order normalization operation. $P(\hat{o}_i | \hat{o}_{i-1},\dots,\hat{o}_1,\bs)$ is detailed in Eq. \ref{eq:prob_position}.
For CBoW sentence encoder, the gradients of words in a sentence are the same for the simple average operation. Thus, we only visualize Ptr-Net with CNN and LSTM sentence encoders.

\begin{table*}[t]
\centering
\begin{tabular}{|l|*{7}{c|}}
 \hline
    &\multicolumn{3}{c|}{PM}&\multicolumn{3}{c|}{LSR}&\multirow{2}*{PMR}\\
  \cline{2-7}
  &P&R&F&P&R&F&\\
 \hline
0 noise  &85.62&85.62&85.62&82.15&82.15&82.15&40.00\\
1 noise  &81.82&81.82&81.82&80.47&80.47&80.47&36.62\\
0/1 noise&83.05&83.40&83.22&81.27&81.46&81.36&35.75\\
\hline
\textbf{+beam search}&&&&&&&\\
0 noise&\textbf{85.79}	&\textbf{85.79}	&\textbf{85.79}	&\textbf{82.40}	&\textbf{82.40}	&\textbf{82.40}	&\textbf{40.44}\\
1 noise&82.28	&82.28	&82.28	&80.92	&80.92	&80.92	&37.33\\
0/1 noise&83.44	&84.04	&83.74	&81.68	&81.98	&81.83	&36.75\\
  \hline
 \end{tabular}
\caption{Performances of different configuration of noise on test sets of arXiv dataset using Prt-Net with LSTM sentence encoder.}\label{tab:noisy}
\end{table*}

\begin{figure*}[t] \small
  \centering
  \pgfplotsset{width=0.33\textwidth}
\subfloat[PM]{
  \begin{tikzpicture}
    \begin{axis}[
    xlabel={Beam Size},
    legend entries={0 noise, 1 noise, 0/1 noise},
    mark size=0.8pt,
    ymajorgrids=true,
    grid style=dashed,
    legend pos= south east,
    legend style={font=\tiny,line width=.5pt,mark size=.2pt,
            /tikz/every even column/.append style={column sep=0.5em}},
            smooth,
    ]
    \addplot [black,dashed,mark=star] table [x index=0, y index=1] {PM_oralce_arXiv_LSTM.txt};
    \addplot [blue,dashed,mark=square*] table [x index=0, y index=2] {PM_oralce_arXiv_LSTM.txt};
    \addplot [red,dashed,mark=otimes*] table [x index=0, y index=3] {PM_oralce_arXiv_LSTM.txt};
    \end{axis}
\end{tikzpicture}
}
\hspace{0em}
\subfloat[LSR]{
  \begin{tikzpicture}
    \begin{axis}[
    xlabel={Beam Size},
    legend entries={0 noise, 1 noise, 0/1 noise},
    mark size=0.8pt,
    ymajorgrids=true,
    grid style=dashed,
    legend pos= south east,
    legend style={font=\tiny,line width=.5pt,mark size=.2pt,
            /tikz/every even column/.append style={column sep=0.5em}},
            smooth,
    ]
    \addplot [black,dashed,mark=star] table [x index=0, y index=1] {LSR_oralce_arXiv_LSTM.txt};
    \addplot [blue,dashed,mark=square*] table [x index=0, y index=2] {LSR_oralce_arXiv_LSTM.txt};
    \addplot [red,dashed,mark=otimes*] table [x index=0, y index=3] {LSR_oralce_arXiv_LSTM.txt};
    \end{axis}
\end{tikzpicture}
}
\hspace{0em}
\subfloat[PMR]{
  \begin{tikzpicture}
    \begin{axis}[
    xlabel={Beam Size},
    legend entries={0 noise, 1 noise, 0/1 noise},
    mark size=0.8pt,
    ymajorgrids=true,
    grid style=dashed,
    legend pos= south east,
    legend style={font=\tiny,line width=.5pt,mark size=.2pt,
            /tikz/every even column/.append style={column sep=0.5em}},
            smooth,
    ]
    \addplot [black,dashed,mark=star] table [x index=0, y index=1] {EMR_oralce_arXiv_LSTM.txt};
    \addplot [blue,dashed,mark=square*] table [x index=0, y index=2] {EMR_oralce_arXiv_LSTM.txt};
    \addplot [red,dashed,mark=otimes*] table [x index=0, y index=3] {EMR_oralce_arXiv_LSTM.txt};
    \end{axis}
\end{tikzpicture}
}
\caption{Oracles on different noise configurations on test set of arXiv dataset using LSTM sentence encoder. Data are selected with the beam sizes of 1, 2, 4, 8, 16, 32, 64 respectively.}\label{fig:res_oracle_arxiv}
\end{figure*}
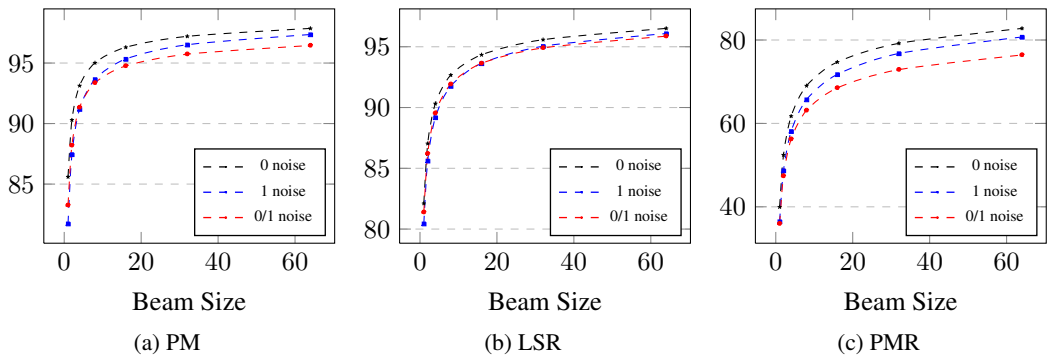

\begin{figure}[t] \small
  \centering
  \pgfplotsset{width=0.33\textwidth}
  \begin{tikzpicture}
    \begin{axis}[
    xlabel={Beam Size},
    legend entries={PMR},
    mark size=0.8pt,
    ymajorgrids=true,
    grid style=dashed,
    legend pos= south east,
    legend style={font=\tiny,line width=.5pt,mark size=.2pt,
            /tikz/every even column/.append style={column sep=0.5em}},
            smooth,
    ]
    \addplot [red,dashed,mark=otimes*] table [x index=0, y index=3] {oralce_sind_LSTM.txt};
    \end{axis}
\end{tikzpicture}
\caption{PMR metrics on test set of SIND dataset using LSTM sentence encoder without noise. Data are selected with the beam sizes of 1, 2, 4, 8, 16, 32, 64 respectively.}\label{fig:res_oracle_sind}
\end{figure}
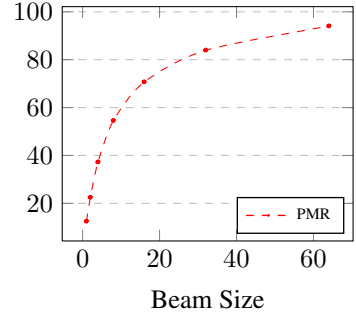

\subsection{Noisy Sentences}
Interestingly, we find our model could deal with the case well when additional noisy sentence exists. To evaluate the performance of our model on noisy sentences, we compare three different strategies in adding noises: (1) 0 noisy sentence (0 noise); (2) 1 noisy sentence (1 noise); (3) 1 noisy sentence with 50\% probability (0/1 noise). All noisy sentences come from own datasets. Since all texts in SIND dataset contain 5 sentences, it is easy for model to tell if there is noisy sentence. Thus, we only evaluate our model on arXiv dataset as shown in Table \ref{tab:noisy}. As mentioned above, we only use the model with LSTM sentence encoder to evaluate the ability of our model on disambiguating noisy sentences, since it always performs better than the one with CBoW or CNN sentence encoder.

Although 0 noise version seems the same as the case in Table \ref{tab:res_all}, they are actually different with each other. In this section, 0 noise version do not constrain the length of predicted sequence to be the same with input. In addition, the $P,R,F$ values of PM and LSR of 0 noise and 1 noise version are not the same actually, and very little difference exists (4 numbers after decimal point). It is because that our model is very good at eliminating the noise terms on 0 noise and 1 noise cases.

As we can see, our model performs best on 0 noise case, which implies that it is easier than 1 noise and 0/1 noise cases for our model. However, it is hard to say which one is more difficult among 1 noise and 0/1 noise cases, as the performance on different metrics are not consistent.

Similarly, the performance of our model is boosted by using beam search instead of greedy decoding.

\subsection{Potential Oracle}
In this section, we evaluate the potential of our model by figuring out what we find in the beam. With beam search strategy, we additionally obtain candidates in beam where the ground truth might appear. 

According to the further analysis of candidates in beam, we find our model could hit ground truth with a very high probability. As show in Figure \ref{fig:res_oracle_arxiv}, our model could obtain 69.03\% in PMR with beam size of 8 on test set of arXiv dataset without noisy sentences, and performance is further boosted with the larger beam size (82.78\% in PMR with beam size of 64 on test set of arXiv dataset without noisy sentences). Notably, the performance boosts faster when beam size is smaller, which shows our model could rank ground truth in a top position in beam (the ground truth has higher score). Additionally, according to the results in Figure \ref{fig:res_oracle_arxiv}, we could tell that 0/1 noise case is a more difficult task for our model as the performance are worse than the 1 noise case. Notably, the results of potential oracle under PM and LSR metrics shown in Figure \ref{fig:res_oracle_arxiv} are using F-value. When determining the best case in beam, we take the one with the highest F-value of PM or LSR metrics for each specific case.

Moreover, to identify whether the high performance in potential oracle is cause by the short texts (arXiv dataset has lots of texts with a few sentences) or not, we also evaluate the performance on SIND dataset (whose texts all contain 5 sentences). As shown in Figure \ref{fig:res_oracle_sind}, the performance in PMR metrics on SIND dataset is very similar to that on arXiv dataset, and our model could obtain 94.01\% in PMR with beam size of 64 on test set of SIND dataset without noisy sentences.

Thus, we believe that we could further boost our performance a lot if we design a model to rerank the candidates.

\section{Related Work}

Previous works on sentence ordering mainly focused on the external and downstream applications, such as multi-document summarization and discourse coherence \cite{van1985semantic,grosz1995centering,van1999semantic,elsner2007unified,barzilay2008modeling}.
\newcite{barzilay2002inferring} proposed two naive sentence ordering techniques,  such as majority ordering and chronological ordering, in the context of multi-document summarization.
\newcite{lapata2003probabilistic} proposed  a probabilistic model that assumes the probability of any given sentence is determined by its adjacent sentence and learns constraints on sentence order from a corpus of domain specific texts.
\newcite{okazaki2004improving} improved chronological ordering by resolving antecedent sentences of arranged sentences and combining topical segmentation.
\newcite{bollegala2010bottom} presented a bottom-up approach to arrange sentences extracted for multi-document summarization. 

Recently, increasing number of researches studied sentence ordering using neural models \cite{agrawal2016sort,li2016neural,chen2016neural}.
\newcite{chen2016neural} framed sentence ordering as an isolated task and firstly applied neural methods on sentence ordering. In addition, they designed an interesting task of ordering the coherent sentences from academic abstracts.
\newcite{agrawal2016sort} focused on a very similar ordering task which ranks image-caption pairs, additionally considering the image information.
\newcite{li2016neural} mainly applied neural models to judge if a given text is coherent.


Despite of their success, they are all pair-wise based models, lack of contextual information. Unlike these work, we propose an end-to-end neural model based on Ptr-Net to address sentence ordering problem.
A few days ago, \newcite{logeswaran2016sentence} proposed a similar model based on recurrent neural networks to address sentence ordering problem. However, their work did not consider neither the case that noisy sentences involve nor alternative sentence encoders exist.

\section{Conclusions}

Sentence ordering is an important factor in natural language generation and attracts increasing focus recently. Previous works are mainly based on pair-wise learning framework, which do not take contextual information into consideration and always lead to error propagation for their pipeline learning strategy. In this paper, we propose an end-to-end neural model based on Ptr-Net to address sentence ordering problem. Experimental results show that our model achieves the state-of-the-art performance even using greedy decoding strategy.

In the future, we would like to further improve the performance by reranking the candidates derived by beam search.

\bibliography{emnlp2016}

\begin{thebibliography}{}

\bibitem[\protect\citename{Agrawal \bgroup et al.\egroup
  }2016]{agrawal2016sort}
Harsh Agrawal, Arjun Chandrasekaran, Dhruv Batra, Devi Parikh, and Mohit
  Bansal.
\newblock 2016.
\newblock Sort story: Sorting jumbled images and captions into stories.
\newblock {\em arXiv preprint arXiv:1606.07493}.

\bibitem[\protect\citename{Barzilay and Elhadad}2002]{barzilay2002inferring}
Regina Barzilay and Noemie Elhadad.
\newblock 2002.
\newblock Inferring strategies for sentence ordering in multidocument news
  summarization.
\newblock {\em Journal of Artificial Intelligence Research}, pages 35--55.

\bibitem[\protect\citename{Barzilay and Lapata}2008]{barzilay2008modeling}
Regina Barzilay and Mirella Lapata.
\newblock 2008.
\newblock Modeling local coherence: An entity-based approach.
\newblock {\em Computational Linguistics}, 34(1):1--34.

\bibitem[\protect\citename{Bollegala \bgroup et al.\egroup
  }2010]{bollegala2010bottom}
Danushka Bollegala, Naoaki Okazaki, and Mitsuru Ishizuka.
\newblock 2010.
\newblock A bottom-up approach to sentence ordering for multi-document
  summarization.
\newblock {\em Information processing \& management}, 46(1):89--109.

\bibitem[\protect\citename{Chen \bgroup et al.\egroup }2016]{chen2016neural}
Xinchi Chen, Xipeng Qiu, and Xuanjing Huang.
\newblock 2016.
\newblock Neural sentence ordering.
\newblock {\em arXiv preprint arXiv:1607.06952}.

\bibitem[\protect\citename{Duchi \bgroup et al.\egroup
  }2011]{duchi2011adaptive}
John Duchi, Elad Hazan, and Yoram Singer.
\newblock 2011.
\newblock Adaptive subgradient methods for online learning and stochastic
  optimization.
\newblock {\em The Journal of Machine Learning Research}, 12:2121--2159.

\bibitem[\protect\citename{Elsner \bgroup et al.\egroup
  }2007]{elsner2007unified}
Micha Elsner, Joseph~L Austerweil, and Eugene Charniak.
\newblock 2007.
\newblock A unified local and global model for discourse coherence.
\newblock In {\em HLT-NAACL}, pages 436--443.

\bibitem[\protect\citename{Erhan \bgroup et al.\egroup
  }2009]{erhan2009visualizing}
Dumitru Erhan, Yoshua Bengio, Aaron Courville, and Pascal Vincent.
\newblock 2009.
\newblock Visualizing higher-layer features of a deep network.
\newblock {\em University of Montreal}, 1341.

\bibitem[\protect\citename{Ferraro \bgroup et al.\egroup
  }2016]{ferraro2016visual}
Francis Ferraro, Nasrin Mostafazadeh, Ishan Misra, Aishwarya Agrawal, Jacob
  Devlin, Ross Girshick, Xiaodong He, Pushmeet Kohli, Dhruv Batra, C~Lawrence
  Zitnick, et~al.
\newblock 2016.
\newblock Visual storytelling.
\newblock {\em arXiv preprint arXiv:1604.03968}.

\bibitem[\protect\citename{Grosz \bgroup et al.\egroup
  }1995]{grosz1995centering}
Barbara~J Grosz, Scott Weinstein, and Aravind~K Joshi.
\newblock 1995.
\newblock Centering: A framework for modeling the local coherence of discourse.
\newblock {\em Computational linguistics}, 21(2):203--225.

\bibitem[\protect\citename{Hochreiter and Schmidhuber}1997]{hochreiter1997long}
Sepp Hochreiter and J{\"u}rgen Schmidhuber.
\newblock 1997.
\newblock Long short-term memory.
\newblock {\em Neural computation}, 9(8):1735--1780.

\bibitem[\protect\citename{Kim \bgroup et al.\egroup }2015]{kim2015character}
Yoon Kim, Yacine Jernite, David Sontag, and Alexander~M Rush.
\newblock 2015.
\newblock Character-aware neural language models.
\newblock {\em arXiv preprint arXiv:1508.06615}.

\bibitem[\protect\citename{Lapata}2003]{lapata2003probabilistic}
Mirella Lapata.
\newblock 2003.
\newblock Probabilistic text structuring: Experiments with sentence ordering.
\newblock In {\em Proceedings of the 41st Annual Meeting on Association for
  Computational Linguistics-Volume 1}, pages 545--552.

\bibitem[\protect\citename{Li and Jurafsky}2016]{li2016neural}
Jiwei Li and Dan Jurafsky.
\newblock 2016.
\newblock Neural net models for open-domain discourse coherence.
\newblock {\em arXiv preprint arXiv:1606.01545}.

\bibitem[\protect\citename{Li \bgroup et al.\egroup }2015]{li2015visualizing}
Jiwei Li, Xinlei Chen, Eduard Hovy, and Dan Jurafsky.
\newblock 2015.
\newblock Visualizing and understanding neural models in nlp.
\newblock {\em arXiv preprint arXiv:1506.01066}.

\bibitem[\protect\citename{Logeswaran \bgroup et al.\egroup
  }2016]{logeswaran2016sentence}
Lajanugen Logeswaran, Honglak Lee, and Dragomir Radev.
\newblock 2016.
\newblock Sentence ordering using recurrent neural networks.
\newblock {\em arXiv preprint arXiv:1611.02654}.

\bibitem[\protect\citename{Mikolov \bgroup et al.\egroup
  }2013]{mikolov2013efficient}
Tomas Mikolov, Kai Chen, Greg Corrado, and Jeffrey Dean.
\newblock 2013.
\newblock Efficient estimation of word representations in vector space.
\newblock {\em arXiv preprint arXiv:1301.3781}.

\bibitem[\protect\citename{Okazaki \bgroup et al.\egroup
  }2004]{okazaki2004improving}
Naoaki Okazaki, Yutaka Matsuo, and Mitsuru Ishizuka.
\newblock 2004.
\newblock Improving chronological sentence ordering by precedence relation.
\newblock In {\em Proceedings of the 20th international conference on
  Computational Linguistics}, page 750.

\bibitem[\protect\citename{Simard \bgroup et al.\egroup }2003]{simard2003best}
Patrice~Y Simard, Dave Steinkraus, and John~C Platt.
\newblock 2003.
\newblock Best practices for convolutional neural networks applied to visual
  document analysis.
\newblock In {\em null}, page 958. IEEE.

\bibitem[\protect\citename{Simonyan \bgroup et al.\egroup
  }2013]{simonyan2013deep}
Karen Simonyan, Andrea Vedaldi, and Andrew Zisserman.
\newblock 2013.
\newblock Deep inside convolutional networks: Visualising image classification
  models and saliency maps.
\newblock {\em arXiv preprint arXiv:1312.6034}.

\bibitem[\protect\citename{Turian \bgroup et al.\egroup }2010]{turian2010word}
Joseph Turian, Lev Ratinov, and Yoshua Bengio.
\newblock 2010.
\newblock Word representations: a simple and general method for semi-supervised
  learning.
\newblock In {\em Proceedings of ACL}.

\bibitem[\protect\citename{Van~Berkum \bgroup et al.\egroup
  }1999]{van1999semantic}
Jos~JA Van~Berkum, Peter Hagoort, and Colin Brown.
\newblock 1999.
\newblock Semantic integration in sentences and discourse: Evidence from the
  n400.
\newblock {\em Cognitive Neuroscience, Journal of}, 11(6):657--671.

\bibitem[\protect\citename{Van~Dijk}1985]{van1985semantic}
Teun~A Van~Dijk.
\newblock 1985.
\newblock Semantic discourse analysis.
\newblock {\em Handbook of discourse analysis}, 2:103--136.

\bibitem[\protect\citename{Vinyals \bgroup et al.\egroup
  }2015]{vinyals2015pointer}
Oriol Vinyals, Meire Fortunato, and Navdeep Jaitly.
\newblock 2015.
\newblock Pointer networks.
\newblock In {\em Advances in Neural Information Processing Systems}, pages
  2692--2700.

\end{thebibliography}
\bibliographystyle{emnlp2016}

\end{document}